\documentclass[letterpaper]{article}
\usepackage{aaai16}
\usepackage{times}
\usepackage{helvet}
\usepackage{courier}
\usepackage{url}
\usepackage{amsmath}
\usepackage{multirow}
\usepackage{array}
\usepackage{graphicx}
\usepackage{subcaption}
\usepackage{CJK}

\newcolumntype{L}[1]{>{\raggedright\arraybackslash}p{#1}}
\newcolumntype{C}[1]{>{\centering\arraybackslash}m{#1}}

\begin{document}
\title{To Swap or Not to Swap? Exploiting Dependency Word Pairs for Reordering in Statistical Machine Translation}
\author{Christian Hadiwinoto$^1$ \hspace{1cm} Yang Liu$^2$ \hspace{1cm} Hwee Tou Ng$^1$ \\
$^1$Department of Computer Science, National University of Singapore \\
\{chrhad,nght\}@comp.nus.edu.sg\\
$^2$Department of Computer Science and Technology, Tsinghua University \\
liuyang2011@tsinghua.edu.cn\\
}
\maketitle
\begin{abstract}
\begin{quote}
Reordering poses a major challenge in machine translation (MT) between two languages with significant differences in word order. In this paper, we present a novel reordering approach utilizing sparse features based on dependency word pairs. Each instance of these features captures whether two words, which are related by a dependency link in the source sentence dependency parse tree, follow the same order or are swapped in the translation output. Experiments on Chinese-to-English translation show a statistically significant improvement of 1.21 BLEU point using our approach, compared to a state-of-the-art statistical MT system that incorporates prior reordering approaches.
\end{quote}
\end{abstract}

\section{Introduction}
\label{sec:intro}

Reordering in machine translation (MT) is a crucial process to get the correct translation output word order given an input source sentence, as word order reflects meaning. It remains a major challenge, especially for language pairs with a significant word order difference. Phrase-based MT systems \cite{koehn_statistical_2003} generally adopt a reordering model that predicts reordering based on the span of a phrase and that of the adjacent phrase \cite{tillmann_unigram_2004,xiong_maximum_2006,galley_simple_2008,cherry_improved_2013}.

The above methods do not explicitly preserve the relationship between words in the source sentence, which reflects the sentence meaning. Word relationship in a sentence can be captured by its dependency parse tree, in which each word $w$ is a tree node connected to its head node $h_w$, another word, indicating that the former is a dependent (child) of the latter.

Dependency parsing has been used for reordering in statistical machine translation (SMT). Its usage is well-known in the pre-ordering approach, where a source sentence is reordered before the actual translation. Dependency-based pre-ordering can be performed either by a rule-based approach based on manually specified human linguistic knowledge \cite{xu_using_2009,cai_dependency-based_2014}, or by a learning approach \cite{xia_improving_2004,habash_syntactic_2007,genzel_automatically_2010,yang_ranking-based_2012,lerner_source-side_2013,jehl_source-side_2014}. Dependency parsing has also been used in reordering approaches integrated with decoding to determine the next source phrase to translate after translating the current source phrase \cite{cherry_cohesive_2008,bach_source-side_2009,chang_discriminative_2009}.

In this paper, we propose a {\it novel} reordering approach integrated
with translation. We propose sparse feature functions based on the
pre-ordering rules of
\cite{cai_dependency-based_2014}. However, in contrast to the manual
rule-based pre-ordering approach of
\cite{cai_dependency-based_2014}, the weights of our sparse feature
functions are automatically learned and used during the actual
translation process, without an explicit pre-ordering step. Our
approach detects and exploits the reordering of each dependency word
pair in the source sentence during phrase-based decoding.

\section{Dependency Word Pair Features}
\label{sec:dependency}

We define a set of sparse features based on dependency tree word pairs to be learned and used in a phrase-based SMT beam search decoding algorithm.

\begin{figure*}[htpb]
\begin{CJK}{UTF8}{gbsn}
\centering
\begin{subfigure}[b]{0.52\textwidth}
\includegraphics[width=\textwidth]{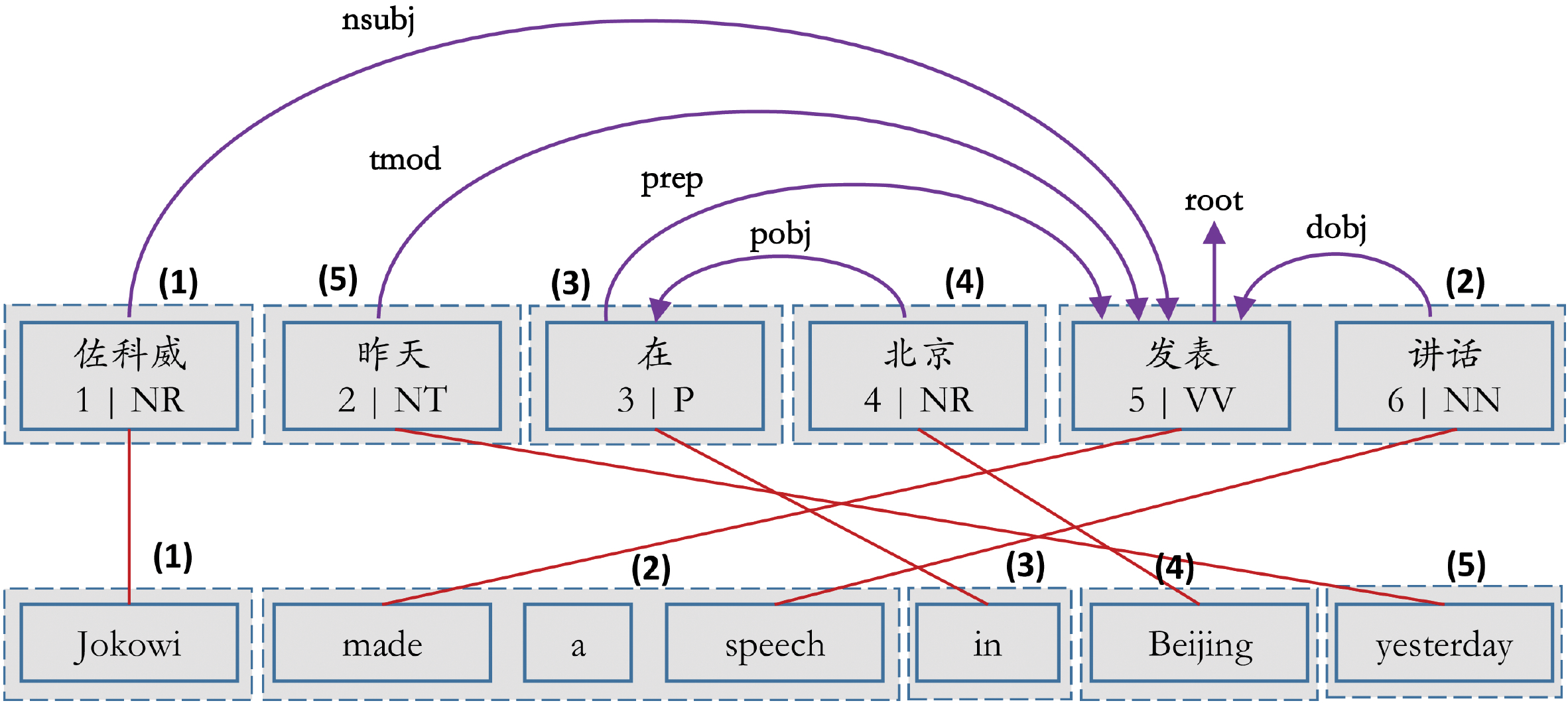}
\caption{~}
\label{fig:illustration_align}
\end{subfigure}
\begin{subfigure}[b]{0.4\textwidth}
\footnotesize
\begin{tabular}{|ll|}
\hline
\multicolumn{2}{|l|}{\bf \underline{(1) 佐科威 $\rightarrow$ Jokowi}} \\
$H_{hc}(\text{发表},\text{佐科威},right,io)$ & $H_{sib}(\text{佐科威},\text{在},io)$ \\
$H_{sib}(\text{佐科威},\text{昨天},io)$ & $H_{sib}(\text{佐科威},\text{讲话},io)$ \\
\hline
\multicolumn{2}{|l|}{\bf \underline{(2) 发表 讲话 $\rightarrow$ made a speech}} \\
$H_{hc}(\text{发表},\text{讲话},left,io)$ & $H_{sib}(\text{昨天},\text{讲话},sw)$  \\
$H_{hc}(\text{发表},\text{昨天},right,sw)$ & $H_{sib}(\text{在},\text{讲话},sw)$ \\
$H_{hc}(\text{发表},\text{在},right,sw)$ & \\
\hline
\multicolumn{2}{|l|}{\bf \underline{(3) 在 $\rightarrow$ in}} \\
$H_{hc}(\text{在},\text{北京},left,io)$ & $H_{sib}(\text{昨天},\text{在},sw)$ \\
\hline
\multicolumn{1}{|l|}{\bf \underline{(4) 北京 $\rightarrow$ Beijing}} & \multicolumn{1}{l|}{\bf \underline{(5) 昨天 $\rightarrow$ yesterday}} \\
\multicolumn{1}{|l|}{(none)} & (none) \\
\hline
\end{tabular}
\caption{~}
\label{fig:illustration_dep}
\end{subfigure}
\caption{\label{fig:illustration} Illustrating example: (a) an aligned
Chinese-English parallel sentence pair with Chinese dependency parse
and a sequence of beam search hypotheses producing phrases
``佐科威 $\rightarrow$ Jokowi'', ``发表 讲话
$\rightarrow$ made a speech'', ``在 $\rightarrow$ in'',
``北京 $\rightarrow$ Beijing'', and ``昨天 $\rightarrow$
yesterday'' (each hypothesis is marked by a grey dashed box and a
sequence number in parentheses) (b) sparse features that are equal to
1 when each hypothesis is generated.}
\end{CJK}
\end{figure*}

\subsection{Capturing Word Pair Ordering in Phrase-Based Beam Search}
\label{sec:dependency_capturing}

The phrase-based approach is a state-of-the-art approach for SMT, taking phrases, defined as a sequence of one or more words, as its translation units. It is performed by a beam search algorithm \cite{koehn_pharaoh:_2004}, in which the search process produces translation from left to right in the translation output. The search is organized into hypotheses, each of which represents an input sentence phrase covered and its possible translation.

As the beam search can choose input phrases in any order, the target-language phrase sequence in the translation output may not follow the original source sentence order. The sequence determines the translation output order and enables translation output reordering for language pairs with differences in word order, such as Chinese and English.

When a word $f_i$ in a source sentence ${\bf f}=f_{1}^{N}$ is covered
by a hypothesis, it is known that the words $\{f_{l}|f_{l} \in
f_{1}^{i-1} \wedge \neg{translated(f_l)}\}$ on the left of $f_i$ and
the words $\{f_{r}|f_{r} \in f_{i+1}^{N} \wedge
\neg{translated(f_r)}\}$ on the right of $f_i$ that have not been
translated will be translated after (appearing on the right of) the translation of $f_i$. As $f_l$ is before $f_i$ in the source sentence, but the translation of $f_l$ is after the translation of $f_i$, the translations of $f_i$ and $f_l$ are \textit{swapped}. Meanwhile, $f_r$ is after $f_i$ both in the source sentence and in the translation, therefore the translations of $f_i$ and $f_r$ are \textit{in-order}. Internally within a phrase, the ordering of each of its words in the translation depends on the phrasal word alignment, which is stored in the phrase table.

As each word in the source sentence is a node of the source dependency parse tree, the above notion can be used for reordering based on the source dependency parse tree. Instead of capturing all pairwise relations, we are only interested in the relations between a word and its related words, defined collectively as its head, sibling, and child words in the dependency parse tree.

\subsection{Dependency Swap Features}
\label{sec:dependency_swap}

We define our dependency swap features, following the rule template definition for dependency swap rules \cite{cai_dependency-based_2014}, which defines rule instances based on the word pairs with head-child or sibling relationship. However, the difference is that our approach does not require manually specifying which dependency labels are the conditions to swap words, but learns them automatically.

In our approach, each rule instance based on the above template
becomes a Boolean sparse feature function
\cite{chiang_11001_2009}. The function parameters are the word pair
specification and output order. While
\citeauthor{cai_dependency-based_2014}
\shortcite{cai_dependency-based_2014} defined the rules only by the
dependency labels, we define our feature functions for each word pair
by the dependency labels, the POS tags, and the combination of both,
resulting in a group of four feature functions for every word pair
ordering. The dependency link label of a word $x$ is defined as the
label of the link connecting $x$ to its head word. Henceforth for each
word $x$, $L(x)$ and $T(x)$ denote the dependency link label and POS tag of $x$ respectively.

Following the dependency swap rule template, we define two types of feature function templates, namely head-child and sibling. The head-child feature functions are equal to 1 if a head word $x_h$ and its child word $x_c$ (where $x_h$ is on the $p\in\{left,right\}$ of $x_c$ in the source sentence) take a certain ordering $o$ (which can be in-order ($io$) or swapped ($sw$)) in the translation output, and 0 otherwise. This group of feature functions is defined as
\begin{equation}
H_{hc}(x_h,x_c,p,o) =
\left[\begin{array}{c}
h_{hc}(L(x_h),L(x_c),p,o) \\
h_{hc}(T(x_h),T(x_c),p,o) \\
h_{hc}(L(x_h),T(x_c),p,o) \\
h_{hc}(T(x_h),L(x_c),p,o)
\end{array}\right]
\label{eq:temp_parchild}
\end{equation}

Similarly, the sibling feature functions are equal to 1 if siblings
$x_l$ and $x_r$ ($x_l$ on the left of $x_r$ in the source sentence) take a certain ordering
$o$ in the translation output, and 0 otherwise. This group of feature functions is defined as
\begin{equation}
H_{sib}(x_l,x_r,o) =
\left[\begin{array}{c}
h_{sib}(L(x_l),L(x_r),o) \\
h_{sib}(T(x_l),T(x_r),o) \\
h_{sib}(L(x_l),T(x_r),o) \\
h_{sib}(T(x_l),L(x_r),o)
\end{array}\right]
\label{eq:temp_sibling}
\end{equation}

\begin{CJK}{UTF8}{gbsn}
Each dependency swap feature that is set to 1 when a hypothesis is
generated captures two source words, one is covered by the current
hypothesis, while the other has not yet been translated. These word
pairs have either head-child or sibling relationship as defined
above. Figure \ref{fig:illustration} is an illustration of how
word order is detected during beam search decoding and the swap
features involved. When ``发表 讲话 $\rightarrow$ made a
speech'' is translated after ``佐科威 $\rightarrow$
Jokowi'', it is known that the head ``发表'' is before the
child ``讲话'' ($p=left$) and their translations also follow the
same word order, setting the value of the following four features to 1:
\[H_{hc}(\text{发表},\text{讲话},left,io) = \left[\begin{array}{c}
h_{hc}(root,dobj,left,io) \\
h_{hc}(VV,NN,left,io) \\
h_{hc}(root,NN,left,io) \\
h_{hc}(VV,dobj,left,io)
\end{array}\right]\]

On the other hand, the head ``发表'' is after the child ``在'' in the source sentence ($p=right$), but ``在'' has not been translated. Therefore, the translation of ``发表'' will be swapped with that of ``在'', setting the four features in $H_{hc}(\text{发表},\text{在},right,sw)$ to 1. Similarly, ``讲话'' is after its sibling ``在'' in the source sentence, but ``在'' has not been translated, resulting in the translation of the two words being swapped and setting the four features in $H_{sib}(\text{在},\text{讲话},sw)$ to 1. The features for ``佐科威'' such as $H_{hc}(\text{发表},\text{佐科威},right,io)$ and $H_{sib}(\text{佐科威},\text{讲话},io)$ are not set to 1 when the hypothesis ``发表 讲话 $\rightarrow$ made a speech'' is generated. Those features were set to 1 with the previous hypothesis ``佐科威 $\rightarrow$ Jokowi''.
\end{CJK}

\subsection{Dependency Distortion Penalty}
\label{sec:dependency_distortion}

To encourage the translation output that conforms to the dependency parse structure, we impose a penalty feature that discourages translation candidates not conforming to the dependency parse subtree \cite{cherry_cohesive_2008}. This assigns a penalty if the translation of the current phrase results in a source dependency parse subtree to be split in the output translation.

\begin{CJK}{UTF8}{gbsn}
The translation output shown in Figure \ref{fig:illustration_align} has each dependency parse subtree grouped together, therefore incurring no penalty. However, there can be a case when the translation has produced ``Jokowi made a speech in'', then translate ``昨天'' to ``yesterday''. This translation will break the cohesion of the source dependency parse subtree ``在 北京'' and is bad. This is the case when a dependency distortion penalty is incurred.
\end{CJK}

\subsection{Modified Future Cost}
\label{sec:dependency_future}

Phrase-based SMT beam search decoding involves future cost, not accumulated over the translation hypotheses but used to reduce search space \cite{koehn_pharaoh:_2004}. This predicts the cost to translate the remaining untranslated phrases.

Based on our dependency swap features, we incorporate the future cost for each untranslated dependency word pairs. The future cost assumes that the untranslated words will be ordered in the most likely ordering. However, if an untranslated word $x$ is an ancestor (in the dependency parse tree) of a word covered by the current hypothesis, the future cost assumes that $x$ precedes all of its related words. This follows the assumption that each subtree will have a contiguous translation \cite{cherry_cohesive_2008}.

\section{Other Sparse Features}
\label{sec:other}

This section describes other sparse features from previous work to compare to our method.

\subsection{Sparse Reordering Orientation Features}
\label{sec:other_sro}

We incorporate sparse reordering orientation features \cite{cherry_improved_2013}. The features are derived from the reordering orientation model, capturing the source position of the current phrase being translated with respect to the previously translated phrase(s), for which three orientation types are defined:
\begin{itemize}
\item Monotone (M), if the current phrase and the previously translated unit are adjacent in the input sentence and the former follows the latter in it, e.g., ``Beijing'' with respect to the previous phrase ``in'' in Figure \ref{fig:illustration}.
\item Swapped (S), if the current phrase and the previously translated unit are adjacent in the input sentence but the former precedes the latter in it (example below).
\item Discontinuous (D), if the current phrase and the previously translated unit are not adjacent in the input sentence, e.g., ``made a speech'' with respect to the previous phrase ``Jokowi'' in Figure \ref{fig:illustration}.
\end{itemize}

\begin{CJK}{UTF8}{gbsn}
\citeauthor{cherry_improved_2013} \shortcite{cherry_improved_2013} designed his sparse reordering orientation model following the hierarchical reordering (HR) model \cite{galley_simple_2008}, capturing the relative position of the current phrase (covered by the current hypothesis) with respect to the largest chunk of contiguous source phrases that form a contiguous translation before this phrase. Therefore, in Figure \ref{fig:illustration_align}, when the decoding produces a phrase ``yesterday'' after ``Beijing'', the orientation of ``yesterday'' is swapped instead of discontinuous, as ``made a speech in Beijing'' is formed by contiguous phrases ``在 北京 发表 讲话'', which is adjacent to ``昨天''.
\end{CJK}

While the original (non-sparse) reordering orientation model is based on the phrase orientation probability in the parallel training data and defines a single feature function on it, the sparse model defines one feature function for each reordering phenomenon during decoding. We define the sparse feature functions taking into account the phrase orientation $o\in\{M,S,D\}$ during decoding and important locations $loc$ pertaining to the current phrase and the previous phrase with the following template:
\begin{equation}
h_{s\_hr}(loc:=rep(loc),o)
\label{eq:temp_sparsereo}
\end{equation}
where locations $loc$ are the first and last words of the current source phrase ($s_{first}$,$s_{last}$), the previously translated unit ($p_{first}$,$p_{last}$), i.e., the largest contiguous chunk of phrases forming a contiguous translation, and the span between the current and the previous source phrase, or gap ($g_{first}$,$g_{last}$) only for discontinuous orientation. Each word in $loc$ is represented, $rep(loc)$, by its POS tag\footnote{\citeauthor{cherry_improved_2013} \shortcite{cherry_improved_2013} substituted POS tags with \emph{mkcls} unsupervised word clusters.}, and the surface lexical form if it belongs to the 80-most frequent words in the training data.

\begin{CJK}{UTF8}{gbsn}
Assuming that only ``在'' belongs to the top-80 words, when the phrase ``yesterday'' is generated after ``Beijing'', the sparse reordering orientation features that are equal to 1 are shown in Figure \ref{fig:sroexample}.
\end{CJK}

\begin{figure}[htpb]
\begin{CJK}{UTF8}{gbsn}
\centering
\small
\begin{tabular}{|ll|}
\hline
$h_{s\_hr}(s_{first}:=NT,S)$ & $h_{s\_hr}(p_{first}:=P,S)$\\
$h_{s\_hr}(s_{last}:=NT,S)$ & $h_{s\_hr}(p_{last}:=NN,S)$\\
 & $h_{s\_hr}(p_{first}:=\text{在},S)$\\
\hline
\end{tabular}
\caption{\label{fig:sroexample} Sparse reordering orientation (HR) features that are equal to 1 when decoding in Fig. \ref{fig:illustration_align} generates the phrase ``昨天 $\rightarrow$ yesterday'' after ``北京 $\rightarrow$ Beijing''}
\end{CJK}
\end{figure}

While the approach leverages the reordering orientation model by defining features on the context information, not just the current phrase, it does not capture dependency relation, by which the important relation between words in the source sentence is captured. Therefore, we introduce our sparse dependency swap features to enable the translation system to arrive at reordering decisions based on the source dependency relations.

\subsection{Dependency Path Features}
\label{sec:other_path}

We also utilize dependency path features \cite{chang_discriminative_2009} for phrase-based SMT, defined over the shortest path of dependency parse tree links bridging the current and the previous source phrase. \citeauthor{chang_discriminative_2009} \shortcite{chang_discriminative_2009} defined the dependency path features on a maximum entropy phrase orientation classifier, trained on their word-aligned parallel text and labeled by the two possible phrase orderings in the translation output: in-order and swapped. Meanwhile, we use the features as sparse decoding features, with the following template:
\begin{equation}
h_{path}(shortest\_path(p_{last},s_{first});o)
\label{eq:temp_path}
\end{equation}
where $o \in \{in\_order,swapped\}$ denotes the orientation of the two phrases in the translation.

Given a source sentence and its translation output, a path is defined between the last word of the previous source phrase $p_{last}$ and the first word of the current source phrase $s_{first}$. Path traversal is always from left to right based on the source sentence word position. Therefore, if the current source phrase being translated is to the right of the previous source phrase, the traversal is from $p_{last}$ to $s_{first}$. Otherwise, if the current source phrase is to the left of the previous source phrase, it is from $s_{first}$ to $p_{last}$. In addition, path edges going against the dependency label arrow are distinguished from those following the arrow.

\begin{CJK}{UTF8}{gbsn}
As an example, in Figure \ref{fig:illustration_align}, when the translation generates ``发表 讲话 $\rightarrow$ made a speech'' after ``佐科威 $\rightarrow$ Jokowi'', the path is from ``佐科威'' to ``发表'', consisting of a direct link following the arrow of $nsubj$, resulting in the feature $h_{path}(nsubj,in\_order)=1$. However, when the translation generates ``昨天 $\rightarrow$ yesterday'' after ``北京 $\rightarrow$ Beijing'', as ``昨天'' is before ``北京'' in the source sentence, the traversal is from ``昨天'' to ``北京'', consisting of the link sequence $tmod,prepR,pobjR$. As it goes against the arrows of $prep$ and $pobj$, the suffix $R$ is added to distinguish it. This results in the feature $h_{path}(tmod,prepR,pobjR;in\_order)=1$.
\end{CJK}

The approach leverages the phrase-based reordering by guiding the ordering of two adjacent phrases using dependency parse. However, the features do not capture the pairwise ordering of every word with its related word, as the features are induced only when the words are used in the two adjacent translation phrases.

\section{Experimental Setup}
\label{sec:experiment}

\subsection{Data Set and Toolkits}
\label{sec:experiment_data}

We built a phrase-based Chinese-to-English SMT system by using Moses \cite{koehn_moses:_2007}. Our parallel training text is a collection of parallel corpora from LDC, which we divide into older corpora\footnote{LDC2002E18, LDC2003E14, LDC2004E12, LDC2004T08, LDC2005T06, and LDC2005T10.} and newer corpora\footnote{LDC2007T23,  LDC2008T06, LDC2008T08, LDC2008T18, LDC2009T02, LDC2009T06, LDC2009T15, LDC2010T03, LDC2013T11, LDC2013T16, LDC2014T04, LDC2014T11, LDC2014T15, LDC2014T20, and LDC2014T26.}. Due to the dominant older data, we duplicate the newer corpora of various domains by 10 times to achieve better domain balance. To reduce the possibility of alignment errors, parallel sentences in the corpora that are longer than 85 words in either Chinese (after word segmentation) or English are discarded. In the end, the final parallel text consists of around 8.8M sentence pairs, 228M Chinese tokens, and 254M English tokens (a token can be a word or punctuation symbol). We also added two dictionaries\footnote{LDC2002L27 and LDC2005T34.} by concatenating them to our training parallel text. The total number of words in these two corpora is 1.81M for Chinese and 2.03M for English.

All Chinese sentences in the training, development, and test data are first word-segmented using a maximum entropy-based Chinese word segmenter \cite{low_maximum_2005} trained on the Chinese Treebank (CTB) scheme. Then the parallel corpus is word-aligned by GIZA++ \cite{och_systematic_2003} using IBM Models 1, 3, and 4 \cite{brown_mathematics_1993}\footnote{The default when running GIZA++ with Moses.}. For building the phrase table, which follows word alignment, the maximum length of a phrase pair is set to 7 words for both the source and target sides.

The language model (LM) is a 5-gram model trained on the English side of the FBIS parallel corpus (LDC2003E14) and the monolingual corpus English Gigaword version 4 (LDC2009T13), consisting of 107M sentences and 3.8G tokens altogether. Each individual Gigaword sub-corpus\footnote{AFP, APW, CNA, LTW, NYT, and Xinhua.} is used to train a separate language model and so is the English side of FBIS. These individual language models are then interpolated to build one single large LM, via perplexity tuning on the development set.

Our translation development set is MTC corpus version 1 (LDC2002T01) and version 3 (LDC2004T07).  This development set has 1,928 sentence pairs in total, 49K Chinese tokens and 58K English tokens on average across the four reference translations. Weight tuning is done by using the pairwise ranked optimization (PRO) algorithm \cite{hopkins_tuning_2011}, which is also used to obtain weights of the sparse features to help determine the reordering.

For dependency sparse features, we parse the Chinese side of our development and test sets by the Mate parser, which jointly performs POS tagging and dependency parsing \cite{bohnet_transition-based_2012}, trained on Chinese Treebank (CTB) version 8.0 (LDC2013T21).

Our test set consists of the NIST MT evaluation sets from 2002 to 2006 and 2008 (LDC2010T10, LDC2010T11, LDC2010T12, LDC2010T14, LDC2010T17, LDC2010T21).

\begin{table*}[htb]
\small
\centering
\begin{tabular}{|l|l|l|l|l|l|l|l|}
\hline
\multicolumn{1}{|c|}{\multirow{2}{*}{\bf Dataset}} & \multicolumn{1}{c|}{\multirow{2}{*}{\bf Base}} & \multicolumn{1}{c|}{\bf \cite{cherry_cohesive_2008}} & \multicolumn{1}{c|}{\bf \cite{chang_discriminative_2009}} & \multicolumn{1}{c|}{\multirow{2}{*}{\bf +DDP+Path}} & \multicolumn{1}{c|}{\bf \cite{cherry_improved_2013}} & \multicolumn{2}{c|}{\bf Ours} \\\cline{3-4}\cline{6-8}
 & & \multicolumn{1}{c|}{\bf +DDP} & \multicolumn{1}{c|}{\bf +Path} & & \multicolumn{1}{c|}{\bf +SHR} & \multicolumn{1}{c|}{\bf +DDP+DS} & \multicolumn{1}{c|}{\bf +DDP+Path+DS} \\
\hline\hline
Devset & 40.04 & 39.55 & 40.51 & 40.32 & 40.92 & 41.39 & 41.61\\\hline\hline
NIST02 & 39.19 & 39.06 & 39.39 & 39.81$^{**\dagger\dagger}$ & 40.08$^{**}$ & 40.48$^{**\dagger\dagger}$ & 40.55$^{**\dagger\dagger}$\\\hline
NIST03 & 39.44 & 40.09$^{**}$ & 40.17$^{**}$ & 39.95$^{**}$ & 39.81 & 40.88$^{**\dagger\dagger}$ & 40.73$^{**\dagger}$\\\hline
NIST04 & 40.26 & 40.16 & 40.62$^{**}$ & 40.63$^{**\dagger\dagger}$ & 40.39 & 40.97$^{**\dagger\dagger}$ & 41.04$^{**\dagger\dagger}$\\\hline
NIST05 & 39.65 & 39.66 & 39.94$^{*}$ & 40.02$^{*\dagger}$ & 39.86 & 41.26$^{**\dagger\dagger}$ & 40.98$^{**\dagger\dagger}$\\\hline
NIST06 & 38.70 & 38.42 & 38.25$^{**}$ & 38.64 & 38.74 & 39.15$^{*\dagger}$ & 39.54$^{**\dagger\dagger}$\\\hline
NIST08 & 30.11 & 30.91$^{**}$ & 30.03 & 30.88$^{**}$ & 30.56$^{*}$ & 31.12$^{**}$ & 31.76$^{**\dagger\dagger}$\\\hline\hline
Average & 37.89 & 38.05 & 38.07$^{**}$ & 38.32$^{**\dagger\dagger}$ & 38.24$^{**}$ & 38.98$^{**\dagger\dagger}$ & {\bf 39.10$^{**\dagger\dagger}$}\\\hline
\end{tabular}
\caption{\label{tab:mainresult} The results of our reordering approach using sparse dependency swap ({\bf DS}) features, in BLEU scores (\%) compared to the baseline ({\bf Base}), on which features are added ($*$: significant at $p < 0.05$; $**$: significant at $p < 0.01$). We also show the results of prior reordering methods, i.e., dependency distortion penalty ({\bf +DDP}) \cite{cherry_cohesive_2008}, sparse dependency path features ({\bf +Path}) \cite{chang_discriminative_2009}, the combination of both ({\bf +DDP+Path}), as well as sparse reordering orientation features ({\bf +SHR}) \cite{cherry_improved_2013}. For all systems involving {\bf DDP} and other features, comparison is also made to the system with only {\bf DDP}. ($\dagger$: significant at $p < 0.05$; $\dagger\dagger$: significant at $p < 0.01$). {\em Note}:  All systems involving DS always incorporate {\bf DDP}.}
\end{table*}

\subsection{Baseline System}
\label{sec:experiment_base}
\begin{CJK}{UTF8}{gbsn}
We build a phrase-based baseline SMT system, which uses non-sparse phrase-based lexicalized reordering (PBLR), in which the reordering probability depends on the phrase being translated and its position with respect to the source position of the previously translated phrase \cite{tillmann_unigram_2004,koehn_edinburgh_2005}, and non-sparse hierarchical reordering (HR), in which the previous unit is not only the previous phrase, but the largest chunk of contiguous source phrases having contiguous translation \cite{galley_simple_2008}. In addition, a distortion limit is set such that the reordering cannot be longer than a certain distance. We set punctuation symbols as reordering constraint across which phrases cannot be reordered, as they form the natural boundaries between different clauses. We also use n-best Minimum Bayes Risk (MBR) decoding \cite{kumar_minimum_2004} instead of the default maximum a-posteriori (MAP) decoding.
\end{CJK}

\subsection{Our Approach}
\label{sec:experiment_our}

To accommodate our sparse feature functions, our Moses code has been modified to read dependency-parsed input sentences and incorporate additional decoding features on top of our baseline, namely dependency distortion penalty (DDP) \cite{cherry_cohesive_2008}, sparse dependency path features (Path) \cite{chang_discriminative_2009}, sparse reordering orientation following hierarchical reordering orientation (SHR) \cite{cherry_improved_2013}, and our sparse dependency swap features (DS). DDP feature is a single penalty feature similar to the distortion penalty for distance-based reordering model \cite{koehn_statistical_2003}, while Path, SHR, and DS are sparse features, each instance of which captures a specific phenomenon during translation \cite{chiang_11001_2009}.

We always couple DS with DDP. However, as the original Path features did not use DDP, we experiment with Path features in one setting that does not incorporate DDP and another that does. The SHR features are not coupled with DDP, following the original design, as the features are not defined on a dependency-parsed input sentence.

\section{Experimental Results}
\label{sec:results}

The translation quality of the system outputs is measured by case-insensitive BLEU \cite{papineni_bleu:_2002}, for which the brevity penalty is computed based on the shortest reference (NIST-BLEU)\footnote{\url{ftp://jaguar.ncsl.nist.gov/mt/resources/mteval-v11b.pl}}. Statistical significance testing between systems is conducted by bootstrap resampling \cite{koehn_statistical_2004}.

Table \ref{tab:mainresult} shows the experimental results. The distortion limit of all the systems is set to 14, which yields the best result on the development set for the baseline system. As shown in the table, the system with our DS features and DDP on top of the baseline is able to improve over the baseline system without and with DDP, by +1.09 and +0.93 BLEU points respectively. The individual contribution of the other dependency-based features (Path), without or with DDP, is inferior to our DS features. Nevertheless, coupling our DS features with Path features yields the best result (+1.21 and +1.05 BLEU points over the baseline without and with DDP).

The SHR features yield more improvement than Path features without DDP, and are comparable to the Path features with DDP. However, our DS features yield more improvement than SHR features. Our preliminary experiments indicate that adding these features on top of the system with DS does not improve over it.

\section{Discussions}
\label{sec:discussions}

The reordering orientation models, i.e., PBLR and HR, only take into account the phrase pair generated by a hypothesis and not the related word properties. Sparse reordering orientation features \cite{cherry_improved_2013} leverage this by capturing the previous phrase (or contiguous chunk) properties. Therefore, they are able to improve over the baseline. However, as the results suggest, dependency parse provides a more useful guidance to reorder a source sentence.

\begin{figure}[ht]
\small
\begin{CJK}{UTF8}{gbsn}
\begin{tabular}{|L{0.45\textwidth}|}
\hline
{\bf \em Source} \newline
广东省\hspace{0.1cm}高新\hspace{0.1cm}技术\hspace{0.1cm}产品\hspace{0.1cm}出口\hspace{0.1cm}的\hspace{0.1cm}主要\hspace{0.1cm}市场\hspace{0.1cm}是\hspace{0.1cm}香港\hspace{0.1cm}、\hspace{0.1cm}美国\hspace{0.1cm}、\hspace{0.1cm}欧盟\hspace{0.1cm}和\hspace{0.1cm}日本\hspace{0.1cm}。\\
\hline

{\bf \em Reference}\newline
The main markets for Guangdong's high-tech products are in Hong Kong, the United States, the European Union and Japan.\\
\hline

{\bf \em Base, +DDP \cite{cherry_cohesive_2008}, +Path \cite{chang_discriminative_2009}: identical output}\newline
The export of high-tech products in Guangdong Province is the main market for Hong Kong, the United States, the European Union and Japan.\\
\hline

{\bf \em +DDP+Path}\newline
The export of high-tech products in Guangdong Province, the main market for Hong Kong, the United States, the European Union and Japan. \\
\hline

{\bf \em +SHR \cite{cherry_improved_2013}}\newline
Guangdong exports of high-tech products is the main market for Hong Kong, the United States, the European Union, and Japan. \\
\hline

{\bf \em +DDP+DS (ours)}\newline
The main market for the export of high-tech products in Guangdong Province are Hong Kong, the United States, the European Union and Japan. \\
\hline

{\bf \em +DDP+Path+DS (ours)}\newline
The main market of the export of high-tech products in Guangdong, Hong Kong, the United States, the European Union and Japan. \\
\hline

\end{tabular}
\end{CJK}
\caption{\label{fig:comparison} Translation output comparison between our baseline ({\bf Base}), prior reordering work ({\bf +DDP}, {\bf +Path}, {\bf +DDP+Path}, and {\bf +SHR}), and our dependency-swap-driven reordering approach ({\bf +DDP+DS} and {\bf +DDP+Path+DS}). }
\end{figure}

As shown in Figure \ref{fig:comparison}, the baseline phrase-based SMT system with the two reordering orientation models (PBLR and HR) produces an incorrect translation output that ``the export is the main market'', which is not what the source sentence means. This is also the case with the system added with prior reordering approaches, which includes DDP \cite{cherry_cohesive_2008}, Path \cite{chang_discriminative_2009}, their combination DDP+Path, and SHR \cite{cherry_improved_2013}. Meanwhile, our reordering approach with DS features is able to output the correct translation ``the main markets for the export'', as it penalizes the swap between the subject head ``market'' and the copula, which should not be swapped.

The dependency parse of a sentence can capture the relationship among the words (nodes) in it. Between each related word pair, there is a dependency label which specifies the head-modifier relationship. While the head-modifier relationships in a sentence hold across languages, their ordering may differ. For example, in Chinese, prepositional phrase (modifier) comes before the predicate (head) verb, while in English, they come in the other way round. This particular clue, provided by the source dependency parse, is useful in deciding the word order in the translation output, corresponding to a word in the source sentence.

Combining our sparse dependency swap features with sparse dependency path features \cite{chang_discriminative_2009} achieves the best experimental result. This can be attributed to the complementary nature of both types of sparse features. Path features are able to capture the subsequence of translation output phrases, i.e., which phrase follows another but not the relative position (right or left) of a source dependency word with respect to all its related words, while our swap features are designed to capture them.

\section{Related Work}
\label{sec:related}

Source dependency trees have been exploited in phrase-based SMT, by modeling transition among dependency subtrees during translation \cite{bach_source-side_2009}. However, this does not take into account the dependency label and the POS tag. Another work exploits source and target dependency trees for phrase-based MT output reranking \cite{gimpel_phrase_2014}, instead of for translation decoding.

\citeauthor{chang_discriminative_2009} \shortcite{chang_discriminative_2009} introduced dependency path as a soft constraint based on the sequence of source dependency links traversed in phrase-based translation. It is used as features on a maximum entropy phrase orientation classifier, whose probability output is used as a decoding feature function. As the path can be arbitrarily long, it may not be represented sufficiently in the training samples. Our sparse feature definition can alleviate this as features are defined on two words. In addition, capturing word pairs instead of paths enables incorporation of other word properties such as POS tags.

\citeauthor{hunter_exploiting_2010} \shortcite{hunter_exploiting_2010} proposed a probability model to capture the offset of a word with respect to its head position in phrase-based MT. Their model does not take into account two sibling words sharing the same head. They reported negative result.

Meanwhile, \citeauthor{gao_soft_2011} \shortcite{gao_soft_2011} defined soft constraints on hierarchical phrase-based MT, which produce translation by bottom-up constituency parsing algorithm instead of beam search. Their soft constraint is also defined on a maximum entropy classifier instead of sparse features.

Prior work has also used constituency parse to guide reordering, by manually defining pre-ordering rules to reorder input sentences into the target language order before translation \cite{collins_clause_2005,wang_chinese_2007}, or to automatically learn those rules \cite{li_probabilistic_2007,khalilov_syntax-based_2011}. Each constituency parse tree node represents the phrase nesting instead of a word, resulting in a deeper structure, which is generally slower to produce.

\section{Conclusion}
\label{sec:conclusion}

We have presented a reordering approach for phrase-based SMT, guided by sparse dependency swap features. We have contributed a new approach for learning and performing reordering in  phrase-based MT by the incorporation of dependency-based features. From our experiments, we have shown that utilizing source dependency parse for reordering sentences helps to significantly improve translation quality over a phrase-based baseline system with state-of-the-art reordering orientation models.

\section{Acknowledgments}

This research is supported by the Singapore National Research Foundation under its International Research Centre @ Singapore Funding Initiative and administered by the IDM Programme Office.

\bibliographystyle{aaai}
\bibliography{reord}

\end{document}